%% file: main.tex
\documentclass{article} 
\usepackage{times}
\usepackage{comment}
\usepackage[margin=0.9in]{geometry} 
\usepackage[T1]{fontenc}
\usepackage{babel}
\usepackage{xcolor}
\usepackage{natbib}
\usepackage{soul}
\usepackage{subcaption}
\usepackage{hyperref}
\usepackage{url}
\usepackage{amsthm,mathtools,amsmath,amsfonts,dsfont}
\usepackage{graphicx}
\usepackage{booktabs}
\usepackage{multirow}
\usepackage{wrapfig}
\usepackage{arydshln}

\usepackage{mathtools}
\input{header}

\title{Overcoming label shift with target-aware federated learning}
\author{
Edvin Listo Zec\footnotemark[1]\\
RISE Research Institutes of Sweden\\ KTH Royal Institute of Technology\\
\texttt{edvin.listo.zec@ri.se} 
\and
Adam Breitholtz  \thanks{Equal contribution. Order decided by coin toss.} \\
Chalmers University of Technology\\
\& University of Gothenburg \\
\texttt{adambre@chalmers.se}
\and 
Fredrik D. Johansson\thanks{ https://www.healthyai.se/} \\
Chalmers University of Technology\\
\& University of Gothenburg \\
\texttt{fredrik.johansson@chalmers.se} \\
}

\begin{document}

\maketitle

\begin{abstract}
Federated learning enables multiple actors to collaboratively train models without sharing private data.
Existing algorithms are successful and well-justified in this task when the intended \emph{target domain}, where the trained model will be used, shares data distribution with the aggregate of clients, but this is often violated in practice. A common reason is label shift---that the label distributions differ between clients and the target domain. We demonstrate empirically that this can significantly degrade performance. To address this problem, we propose \algname{}, a principled and practical model aggregation scheme that adapts to label shifts \textit{to improve performance in the target domain} by leveraging knowledge of label distributions at the central server. Our approach ensures unbiased updates under federated stochastic gradient descent which yields robust generalization across clients with diverse, label-shifted data. Extensive experiments on image classification tasks demonstrate that \algname{} consistently outperforms baselines by aligning model aggregation with the target domain. Our findings reveal that conventional federated learning methods suffer severely in cases of extreme label sparsity on clients, highlighting the critical need for target-aware aggregation as offered by \algname{}. 
\end{abstract}

\section{Introduction}

Federated learning (FL) has emerged as a powerful paradigm for training machine learning models collaboratively across multiple clients without sharing data~\citep{mcmahan17a,kairouz2021advances}. This is attractive in problems where privacy is paramount, such as healthcare~\citep{sheller2020federated}, finance~\citep{byrd2020differentially}, and natural language processing~\citep{hilmkil2021scaling}. 
While effective when data from different clients are identically distributed, the performance of federated learning can degrade significantly when clients exhibit systematic data heterogeneity, such as label shift~\citep{zhao2018federated,woodworth20}. 

Most federated learning research, even that addressing data heterogeneity, focuses on what we term \textit{standard} federated learning, where the test distribution matches the combined distribution of training clients. However, many real-world applications require generalization to a target client or domain with a distinct, unknown data distribution. Consider a retail scenario: multiple stores (clients) collaboratively train a sales prediction model using their local purchase histories. While each store's data reflects its unique customer base, the goal is to deploy the model in a \textit{new} store (target client) with different customer preferences, and no historical records. This is \textit{target-aware} federated learning, a more challenging paradigm than standard federated learning due to the inherent distributional shift between training and test data.

The problem of generalizing under distributional shifts has been extensively studied in centralized settings, often under the umbrella of domain adaptation~\citep{blanchard2011generalizing,ganin2016domain}. However, traditional domain adaptation techniques, such as sample re-weighting~\citep{lipton2018detecting} or domain-invariant representation learning~\citep{arjovsky_invariant_2020}, typically require access to data from both source and target domains. This requirement is incompatible with the decentralized nature of federated learning, where neither the server nor the clients share data between them. While several techniques have been proposed to address client heterogeneity in standard FL, such as regularization~\citep{li2020federated,li2021ditto}, clustering~\citep{ghosh2020efficient,vardhan2024improved}, and meta-learning~\citep{chen2018federated,jiang2019improving}, they do not address the challenge of generalizing to a target client with differing data distribution.  

\paragraph{Contributions} We introduce the problem of \textit{target-aware} federated learning under label shift, where the goal is to train a model that generalizes well to a target client (domain) whose label distribution differs from those of the training clients (see Section~\ref{sec:problem}). To address this problem, we propose a novel aggregation scheme called \algname{} that optimizes a convex combination of client models to ensure that the aggregated model is better suited for the label distribution of the target domain (Section~\ref{sec:algorithm}). 
We prove that the resulting stochastic gradient update behaves, in expectation, as centralized learning in the target domain (Proposition~\ref{prop:sgd_update}), and examine its relation to standard federated averaging (Proposition~\ref{sec:limits}). We demonstrate the effectiveness of \algname{} through an extensive empirical evaluation (Section~\ref{sec:experiments}), showing that it outperforms traditional approaches in scenarios where distributional shifts pose significant challenges, at the small cost of sharing client label marginals with the central server. Moreover, we observe that traditional methods struggle particularly in scenarios where training clients have sparse label distributions, 
highlighting the need for target-aware aggregation strategies.

%
%
\section{Target-aware federated learning with label shift}
\label{sec:problem}
In federated learning, a global model $h_\theta$ is produced by a central server by aggregating  updates to model parameters $\theta$ from multiple clients~\citep{mcmahan17a}. We focus on classification tasks in which the goal is for $h_\theta$ to predict the most probable label $Y \in \{1, ..., K\}$ for a given $d$-dimensional input $X \in \cX \subset \bbR^d$. Each client $i=1, ..., M$ holds a data set $D_i = \{(x_{i,1},y_{i,1}), ..., (x_{i,n_i}, y_{i,n_i})\}$ of $n_i$ labeled examples, assumed to be drawn i.i.d. from a \emph{local} client-specific distribution $S_i(X,Y)$. Due to constraints on privacy or communication, these data sets cannot be shared directly with other clients or with the central server.

Learning proceeds over rounds $t=1, ..., t_{max}$, each comprising three steps: (1) The central server broadcasts the current global model parameters $\theta_t$ to all clients; (2) Each client $i$ computes updated parameters $\theta_{i,t}$ based on their local data set $D_i$, and sends these updates back to the server; (3) The server aggregates the clients' updates, for example, using federated averaging (FedAvg)~\citep{mcmahan17a} or related techniques, to produce the new global model $\theta_{t+1}$.
%

A common implicit assumption in federated learning is that the learned model will be applied in a target domain $T(X,Y)$ that coincides with the aggregate distribution of clients, 
\begin{equation}\label{eq:aggregate}
\bar{S}(X,Y) = \sum_{i=1}^M \frac{n_i}{N} S_i(X,Y), 
\end{equation}
where $N = \sum_{i=1}^M n_i$. To this end, trained models are evaluated in terms of their average performance over clients. However, in applications, the intended target domain may be different entirely~\citep{baibenchmarking}. Here, we assume that the target domain is distinct from all client distributions: $\forall i : T(X,Y) \neq S_i(X,Y)$ and from the client aggregate, $T(X,Y) \neq \bar{S}(X,Y)$. We refer to this setting as \emph{target-aware federated learning}. While distributional shift between clients is a well-recognized problem in federated learning,  the target domain is still typically the client aggregate $\bar{S}$~\citep{scaffold,fedprox}. Our setting differs also from federated domain generalization which lacks a specific target domain~\citep{baibenchmarking}. 

In target-aware federated learning, the goal is to train a model to predict well in a target domain $T(X,Y)$ \emph{without access to samples from }$T$. Formally, our objective is to minimize the expected target risk, $R_T$ of a classifier $h_\theta : \cX \rightarrow \cY$, with respect to a loss function $\ell : \cY \times \cY \rightarrow \bbR$,
\begin{equation}\label{eq:target_risk}
\underset{\theta}{\mbox{minimize}}\; R_T(h_\theta) \coloneqq \underset{(X,Y)\sim T}{\E}\left[ \ell(h_\theta(X), Y) \right]~.
\end{equation}

To make solving \eqref{eq:target_risk} possible, we assume that target and client label marginal distributions $T(Y), \{S_i(Y)\}$ are known to the central server. This is much less restrictive than it sounds: (i) Estimating each client label distribution $S_i(Y)$ merely involves computing the proportion of each label in the client sample $D_i$, (ii) The target client (domain) may have collected label statistics without logging context features $X$. In our retail example, the label distribution corresponds to the proportion of sales $T(Y=y)$ of each product category $y$, and many companies store this information without logging customer features $X$. 
The central server is given access to all label distributions to facilitate the learning process, but these are \emph{not available to the clients}. Retailers may be hesitant to share their exact sales statistics $T(Y)$ with competitors but could share this information with a neutral third party (central server) responsible for coordinating the federated learning process. There is a privacy-accuracy trade-off in all FL settings. In our experiments, \emph{we show that substantial performance improvements can be gained at the small privacy cost of sharing the label marginals with the central server.} 

As in standard FL, clients $i \neq j$ do not communicate directly with each other directly but interact with the central server through model parameters. 
While it is technically possible for the server to \emph{infer} each client's label distribution $S_i(Y)$ based on their parameter updates~\citep{ramakrishna2022inferring}, doing so would likely be considered a breach of trust in practical applications and sharing would be preferred. 

We assume that the distributional shifts between clients and the target are restricted to \emph{label shift}---while the label distributions vary across clients and the target, the class-conditional input distributions are identical.
\begin{thmasmp}[Label shift]
For the client distributions $S_1, ..., S_M$ and the target distribution $T$, 
\label{asmp:labelshift}
\begin{equation}\label{eq:labelshift}
\forall i,j \in [M] : S_i(X\mid Y) = S_j(X \mid Y) = T(X\mid Y)~. 
\end{equation}
\end{thmasmp}
This setting has been well studied in non-federated learning, see e.g., \citep{lipton2018detecting}. In the retail example, label shift means that the proportion of sales across product categories ($S_i(Y)$ and $T(Y)$) varies between different retailers and the target, but that the pattern of customers who purchase items in each category ($S_i(X \mid Y)$ and $T(X \mid Y)$) remain consistent. In other words, although retailers may sell different quantities of products across categories, the characteristics of customers buying a particular product (conditional on the product category) are assumed to be the same. Note that both label shift and \emph{covariate shift} may hold, that is, there are cases where $\forall i : S_i(X\mid Y) = T(X\mid Y)$ \emph{and} $S_i(Y\mid X) = T(Y\mid X)$, but $S_i(X), T(X)$ differ, such as when the labeling function is deterministic. 

%

\subsection{Limitations of classical aggregation}
When either all clients $\{S_i\}$ or their aggregate $\bar{S}$, see \eqref{eq:aggregate}, are identical in distribution to the target domain, 
the empirical risk on aggregated client data is identical in distribution ({\scriptsize $\overset{d}{=}$}) to the empirical risk of a hypothetical data set $D_T = \{(x_{T,j}, y_{T,j})\}_{j=1}^{n_T}$ drawn from the target domain,
$$
\hat{R} \coloneqq \sum_{i=1}^M\sum_{j=1}^{n_i}  \frac{\ell\left(h(x_{i,j}), y_{i,j}\right)}{N}  \overset{d}{=}  \sum_{j=1}^{n_T} \frac{\ell\left(h(x_{T,j}), y_{T,j}\right) }{n_T} \eqqcolon \hat{R}_T  
$$
Thus, if each client performs a single gradient descent update, the mean of these, weighted by the client sample sizes, is equal in distribution to a centralized batch update for the target domain, given the previous parameter value. 
This property justifies the federated stochastic gradient (FedSGD) and federated averaging principles~\citep{mcmahan17a}, both of which aggregate parameter updates in this way, 
\begin{equation}
\theta^{FA}_{t+1} = \sum_{i=1}^M \alpha^{FA}_i \theta_{i, t} \quad \mbox{where} \quad \alpha^{FA}_i = \frac{n_i}{\sum_{j=1}^M n_j}~.
\end{equation}

Unfortunately, when the target domain $T$ is not the aggregate of clients $\bar{S}$, the aggregate risk gradient $\nabla \hat{R}$ and, therefore, the FedSGD update are no longer unbiased gradients and updates for the risk in the target domain. 
As we see in Table \ref{tab:results}, this can have large effects on model quality.

\textbf{Our central question is:} How can we \emph{aggregate} the parameter updates $\theta_{i,t}$ of the $M$ clients, whose data sets are drawn from distributions $S_1, ..., S_M$, such that the resulting federated learning algorithm minimizes the target risk, $R_T$? 
%

%
%
\section{\algname: Target-aware adjustment for label shift}
\label{sec:algorithm}
\begin{figure}[t!]
    \centering
    \includegraphics[width=.65\columnwidth]{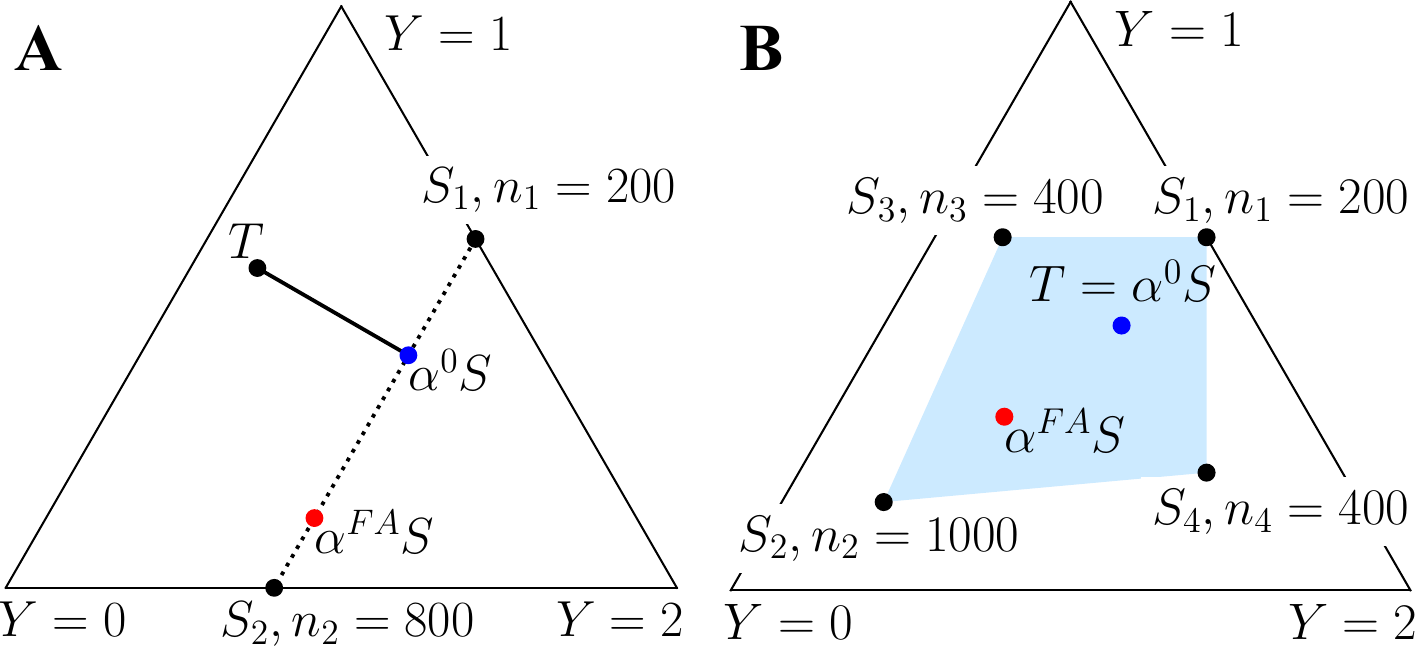}
    \caption{Illustration of the target label marginal $T$ and client marginals $S_1, ..., S_4$ in a ternary classification task, $Y\in \{0,1,2\}$. A: there are fewer clients than labels, $M < K$, and $T \not\in \convh(S)$; $\alpha^0S$ is a projection of $T$ onto $\convh(S)$. B: $T \in \convh(S)$ and coincides with $\alpha^0 S$. In both cases, the label marginal $\alpha^{FA}S$ implied by FedAvg is further from the target distribution.}
    \label{fig:simplex}
\end{figure}%

Next, we develop a model aggregation strategy for target-aware federated learning. Under Assumption~\ref{asmp:labelshift} (label shift), the target risk is a weighted sum of class-conditional client risks,
\begin{align*}
\forall i : R_T(h) 
& = \sum_{y=1}^K T(y) \E_{S_i}[\ell(h(X), y)\mid Y=y]~.
\end{align*}
In centralized learning, this insight is often used to re-weight the training objective in a source domain $S(y)$ by the importance ratio $T(y)/S(y)$~\citep{lipton2018detecting,japkowicz2002class}. \emph{That is not an option here since $T(Y)$ is not revealed to the clients.} 
For now, assume instead that the target label distribution is \emph{covered} by the convex hull of the set of client label distributions $S = \{S_i(Y)\}_{i=1}^M$.
\begin{thmasmp}[Target coverage]\label{asmp:coverage}
     The target label distribution $T(Y)$ is covered by the convex hull of client label distributions $S_1(Y), ..., S_M(Y)$, that is $T \in \convh(S)$, or 
     \begin{equation}\label{eq:conv_comb}
      \exists \alpha^c \in \Delta^{M-1} : T(y) = \sum_{i=1}^M \alpha_i^c S_i(y) \quad \forall y \in [K]~.
     \end{equation}
\end{thmasmp}

Under label shift, Assumption~\ref{asmp:coverage} implies that $T(X,Y) = \sum_{i=1}^M\alpha^c_i S_i(X,Y)$, as well.
Thus, under Assumptions~\ref{asmp:labelshift}--\ref{asmp:coverage}, we have for any $\alpha^c$ satisfying \eqref{eq:conv_comb},
\begin{align}\label{eq:conv_risk}
R_T(h) &= \sum_{y=1}^K \left(\sum_{i=1}^M \alpha^c_i S_i(y) \right) \E[\ell(h(X), y)\mid Y=y] \\
& =\sum_{i=1}^M \alpha^c_i R_{S_i}(h)~.
\end{align}
Consequently, aggregating client updates with weights $\alpha^c$ will be an unbiased estimate of the update.
\begin{thmprop}[Unbiased SGD update]\label{prop:sgd_update}
    Consider a single round $t$ of federated learning in the batch stochastic gradient setting with learning rate $\eta$. Each client $i\in [M]$ is given parameters $\theta_t$ by the server, computes their local gradient, and returns the update $\theta_{i, t} = \theta_t - \eta \nabla_\theta \hat{R}_i(h_{\theta_t})$. Let Assumptions~\ref{asmp:labelshift}--\ref{asmp:coverage} hold and $\alpha^c$ satisfy \eqref{eq:conv_comb}. Then, the aggregate update $\theta_{t+1} = \sum_{i=1}^M \alpha^c_i \theta_{i,t}$ satisfies
    $$
    \E[\theta_{t+1} \mid \theta_t] = \E[\theta_{t+1}^T \mid \theta_t]~, 
    $$
    where $\theta_{t+1}^T = \theta_t - \eta \nabla_\theta \hat{R}_T(h_{\theta_t})$ is the batch stochastic gradient descent (SGD) update for $\hat{R}_T$ that would be obtained with a sample from the target domain. A proof is in Appendix~\ref{app:proofs}.
\end{thmprop}

By Proposition~\ref{prop:sgd_update}, we may compute unbiased parameter updates for the target domain by replacing the aggregation step of FedSGD with aggregation weighted according to  $\alpha^c$. In practice, many federated learning systems, including FedAvg, allow clients several steps of local optimization (e.g., an epoch) before aggregating the parameter updates at the server. Strictly speaking, this is not justified by Proposition~\ref{prop:sgd_update}, but we find in all experiments that aggregating client updates computed over an epoch performs very well, see Section~\ref{sec:experiments}.

In applications, the target may not be covered by clients (Assumption~\ref{asmp:coverage} may not hold), and $\alpha^c$ may not exist. For example, if the target label marginal $T(y)$ is sparse, only clients with \emph{exactly the same sparsity pattern} as $T$ can be used in a convex combination $\alpha^c S = T$. That is, if we aim  to classify images of animals and $T$ contains no tigers, then no clients contributing to the combination can have data containing tigers. Since $\{S_i(Y)\}_{i=1}^M, T(Y)$ are known to the server, it is straightforward to verify Assumption~\ref{asmp:coverage}.

A pragmatic choice when Assumption~\ref{asmp:coverage} is violated is to look for the convex combination $\alpha^0$ that most closely aligns with the target label distribution, and use that for aggregation,
\begin{equation}\label{eq:alpha0}
\alpha^{0} = \argmin_{\alpha \in \Delta^{M-1}} \bigg\|\sum_{i=1}^M \alpha_i S_i(Y) - T(Y) \bigg\|_2^2 \quad 
\end{equation}
We illustrate the label distributions implied by weighting with $\alpha^0$ and $\alpha^{FA}$ (FedAvg) in Figure~\ref{fig:simplex}. 

\paragraph{Effective sample size of aggregates.} A limitation of aggregating using $\alpha^0$ as defined in \eqref{eq:alpha0} is that, unlike FedAvg, it does not give higher weight to clients with larger sample sizes, which can lead to a higher variance in the model estimate. The variance of importance-weighted estimators can be quantified through the concept of \emph{effective sample size} (ESS)~\citep{kong1992note}, which measures the number of samples needed from the target domain to achieve the same variance as a weighted estimate computed from source-domain samples. ESS is often approximated as $1/(\sum_{i=1}^m w_j^2)$ where $w$ are normalized sample weights such that $w_j \geq 0$ and $\sum_{j=1}^n w_j = 1$. In federated learning, we can interpret the aggregation step as assigning a total weight $\alpha_i$ to each client $i$, which has $n_i$ samples. Consequently, each sample $(x_j, y_j) \in D_i$ has the same weight $\tilde{w}_j = \alpha_i/n_i$. The ESS for the aggregate is then given by $1/(\sum_{i=1}^m (\sum_{j \in S_i} \tilde{w}_j^2)) = 1/(\sum_{i=1}^m n_i \alpha_i^2/n_i^2) = 1/(\sum_{i=1}^m \alpha_i^2/n_i)$. 

In light of the above, we propose a client aggregation step such that the weighted sum of clients' label distributions will a) closely align with the target label distribution, and b) minimize the variance due to weighting using the inverse of the ESS. For a given regularization parameter $\lambda \in [0, \infty)$, we define weights $\alpha^\lambda$ as the solution to the following problem
\begin{equation}
    \alpha^\lambda=\argmin_{\alpha \in \Delta^{M-1}} \|T(Y)-\sum_{i=1}^M\alpha_i S_i(Y)\|^2_2+ \lambda \sum_i \frac{\alpha_i^2}{n_i} ~,
    \label{eq:alpha_alg}
\end{equation}
with aggregate client parameters as $\theta^{\lambda}_{t+1} = \sum_{i=1}^M \alpha^{\lambda}_i \theta_{i, t}$. 
We refer to this strategy as \algfullname{} (\algname). 

%
%

\subsection{\algname{} in the limits}
\label{sec:limits}
In the \algname{} aggregation scheme (\eqref{eq:alpha_alg}), there exists a trade-off between closely matching the target label distribution and minimizing the variance of the model parameters. This trade-off gives rise to two notable limit cases:  $T \in \convh(S), \lambda \rightarrow 0$, and $\lambda \rightarrow \infty$. If all source distributions $\{S_i\}_{i=1}^M$ are identical and match the target distribution, this corresponds to the classical i.i.d. setting.


\paragraph{Case 1: $\lambda \rightarrow \infty \Rightarrow$  Federated averaging}
In the limit $\lambda\to \infty$, as the regularization parameter $\lambda$ grows large, \algname{} aggregation approaches FedAvg aggregation.
\begin{thmprop}\label{prop:lambdainf_main}
The limit solution $\alpha^\lambda$ to \eqref{eq:alpha_alg}, as $\lambda \rightarrow \infty$, is
\begin{equation}\label{eq:lambdainfalpha}
\lim_{\lambda\rightarrow \infty} \alpha^\lambda_i = 
\frac{n_i}{\sum_{j=1}^M n_j} = \alpha^{FA}_i \quad \mbox{for} \quad i = 1, \ldots, M~.%
\end{equation}%
\end{thmprop}%
The result is proven in Appendix~\ref{app:proofs}.
By Proposition~\ref{prop:lambdainf_main}, the FedAvg weights $\alpha^{FA}$ minimize the ESS and coincide with \algname{} weights $\alpha^\lambda$ in the limit $\lambda\rightarrow \infty$.
As a rare special case, whenever $T(Y)= \bar{S} = \sum_{i=1}^M \frac{n_i}{N} S_i(Y)$, FedAvg weights $\alpha^{FA} = \alpha^\lambda$ for any value of $\lambda$, since both terms attain their mimima at this point. However, this violates the assumption that $T(Y) \neq \bar{S}(Y)$.

\paragraph{Case 2: Covered target, $T \in \convh(S)$}
Now, consider when the target label distribution is in the convex hull of the source label distributions, $\convh(S)$. Then, we can find a convex combination $\alpha^c$ of source distributions $S_i(Y)$ that recreate $T(Y)$, that is, $T(Y) = \sum_{i=1}^M \alpha_i^c S_i(Y)$. However, when there are more clients than labels, $M > K$, such a \emph{satisfying combination} $\alpha^c$ need not be unique and different combinations may have different effective sample size. Let $A^c = \{\alpha^c \in \Delta^{M-1} : T(Y) = (\alpha^c)^\top S(Y)\}$ denote all satisfying combinations where $S(Y) \in \bbR^{M \times K}$ is the matrix of all client label marginals.
For a sufficiently small regularization penalty $\lambda$, the solution to \eqref{eq:alpha_alg} will be the satisfying combination with largest effective sample size.
$$
\lim_{\lambda\rightarrow 0} \alpha^\lambda = \argmin_{\alpha \in A^c} \sum_{i=1}^M \frac{\alpha_i^2}{n_i}~.
$$
If there are fewer clients than labels, $M < K$, the set of target distributions for which a satisfying combination exists has measure zero, see Figure~\ref{fig:simplex} (left). 
Nevertheless, the two cases above allow us to interpolate between being as faithful as possible to the target label distribution ($\lambda\to 0$) and retaining the largest effective sample size ($\lambda\to \infty$), the latter coinciding with FedAvg. Finally, when $T\in \convh(S)$ and $\lambda \rightarrow 0$, Proposition~\ref{prop:sgd_update} applies also to FedPALS; the aggregation strategy results in an unbiased estimate of the target risk gradient in the SGD setting. However, like the unregularized weights, Proposition~\ref{prop:sgd_update} does not apply for multiple local client updates.


\paragraph{Case 3: $T \not\in \convh(S)$}
If the target distribution does not lie in $\convh(S)$, see Figure~\ref{fig:simplex} (left), \algname{} projects the target to the ``closest point'' in $\convh(S)$ if $\lambda=0$, and to a tradeoff between this projection and the FedAvg aggregation if $\lambda>0$. We have a discussion on how to choose $\lambda$ in the second and third cases in Appendix \ref{app:lambda}.

\paragraph{Sparse clients and targets} In problems with a large number of labels, $K \gg 1$, it is common that any individual domain (clients or target) supports only a subset of the labels. For example, in the IWildCam benchmark, not every wildlife camera captures images of all animal species. 
When the target $T(Y)$ is \emph{sparse}, meaning $T(y)=0$ for certain labels $y$, it becomes easier to find a good match $(\alpha^\lambda)^\top S(Y) \approx T(Y)$ if the client label distributions are also sparse. Achieving a perfect match, i.e., $T \in \convh(S)$, requires that (i) the clients collectively cover all labels in the target, and (ii) each client contains only labels that are present in the target. If this is also beneficial for learning, it would suggest that the client-presence of labels that are not present in the target would \emph{harm} the aggregated model. We study the implications of sparsity of label distributions empirically in Section~\ref{sec:experiments}.

\section{Related work}
Efforts to mitigate the effects of distributional shifts in federated learning can be broadly categorized into client-side and server-side approaches. Client-side methods use techniques such as clustering clients with similar data distributions and training separate models for each cluster~\citep{ghosh2020efficient, sattler2020clustered, vardhan2024improved}, and meta-learning to enable models to quickly adapt to new data distributions with minimal updates~\citep{chen2018federated, jiang2019improving, fallah2020personalized}. Other notable strategies include regularization techniques that penalize large deviations in client updates to ensure stable convergence~\citep{fedprox, li2021ditto} and recent work on optimizing for flatter minima to enhance model robustness~\citep{Qu2022, caldarola2022}.
Server-side methods focus on improving model aggregation or applying post-aggregation adjustments. These include optimizing aggregation weights \citep{fedopt}, learning adaptive weights~\citep{clientcoherence}, iterative moving averages to refine the global model~\citep{Zhou2023UnderstandingAI}, and promoting gradient diversity during updates~\citep{Zeng2023FedAWAREMG}. Both categories of work overlook shifts in the target distribution, leaving this area unexplored.

Another related area is personalized federated learning, which focuses on fine-tuning models to optimize performance on each client’s specific local data~\citep{finetuning_koyejou2022, Boroujeni2024PersonalizedFL}. This setting differs fundamentally from our work, which focuses on improving generalization to new target clients without any training data available for fine-tuning.
Label distribution shifts have also been explored with methods such as logit calibration~\citep{zhang22logitscalibration, Xu2023StabilizingAI}, novel loss functions~\citep{Wang_Xu_Wang_Zhu_2021}, feature augmentation~\citep{Xia2023FLeaIF}, gradient reweighting~\citep{Xiao2023FedGraBFL}, and contrastive learning~\citep{fediic}. However, like methods aimed at mitigating the effects of general shifts, these do not address the challenge of aligning models with an unseen target distribution, as required in our setting. 

Generalization under domain shift in federated learning remains underdeveloped~\citep{baibenchmarking}. The work most similar to ours is that of agnostic federated learning (AFL)~\citep{mohri19a}, which aims to learn a model that performs robustly across all possible target distributions within the convex hull of client distributions. One notable approach is tailored for medical image segmentation, where clients share data in the frequency domain to achieve better generalization across domains~\citep{liu2021feddg}. However, this technique requires data sharing, making it unsuitable for privacy-sensitive applications like ours. A different line of work focuses on addressing covariate shift in federated learning through importance weighting~\citep{ramezani-kebrya2023federated}. Although effective, this method requires sending samples from the test distribution to the server, which violates our privacy constraints. 
\section{Experiments}
\label{sec:experiments}
We perform a series of experiments on benchmark data sets to evaluate \algname{} in comparison with baseline federated learning algorithms. The experiments aim to demonstrate the value of the central server knowing the label distributions of the client and target domains when these differ substantially. Additionally, we seek to understand how the parameter $\lambda$, controlling the trade-off between bias and variance in the \algname{} aggregation scheme, impacts the results. Finally, we investigate how the benefits of \algname{} are affected by the sparsity of label distributions and by the distance $d(T,S) \coloneqq \min_{\alpha \in \Delta^{M-1}}\|T(Y) - \alpha^\top S(Y)\|_2^2$ from the target to the convex hull of clients.

\paragraph{Experimental setup}
While numerous benchmarks exist for federated learning~\citep{caldas2018leaf, chen2022pfl} and domain generalization~\citep{gulrajanisearch, koh2021wilds}, respectively, until recently none have addressed tasks that combine both settings. To fill this gap, \citet{baibenchmarking} introduced a benchmark specifically designed for federated domain generalization (DG), evaluating methods across diverse datasets with varying levels of client heterogeneity. In our experiments, we use the PACS~\cite{pacs} and iWildCAM data sets from the \citet{baibenchmarking} benchmark to model realistic label shifts between the client and target distributions. We modify the PACS dataset to consist of three clients, each missing a label that is present in the other two. Additionally, one client is reduced to one-tenth the size of the others, and the target distribution is made sparse in the same label as that of the smaller client. For further details see Appendix \ref{app:expdetail}. 

Furthermore, we construct two additional tasks by introducing label shift to standard image classification data sets, Fashion-MNIST~\citep{xiao2017fashion} and CIFAR-10~\citep{krizhevsky2009learning}. We apply two label shift sampling strategies: sparsity sampling and Dirichlet sampling. 
Sparsity sampling involves randomly removing a subset of labels from clients and the target domain, following the data set partitioning technique first introduced in~\citet{mcmahan17a}.  Each client is assigned $C$ random labels, with an equal number of samples for each label and no overlap among clients.

\begin{table*}[t]
\centering
\footnotesize
\caption{Comparison of mean accuracy and standard deviation (\(\pm\)) across different algorithms. The reported values are over 8 independent random seeds for the CIFAR-10 and Fashion-MNIST tasks, and 3 for PACS. $C$ indicates the number of labels per client and $\beta$ the Dirichlet concentration parameter. $M$ is the number of clients. The \textit{Oracle} method refers to a FedAvg model trained on clients whose distributions are identical to the target.}
\resizebox{\textwidth}{!}{%
\begin{tabular}{lccccccccc}
\toprule
\textbf{Data set} & \textbf{Label split} &  \textbf{M} &\textbf{\algname{}} & \textbf{FedAvg} & \textbf{FedProx} & \textbf{SCAFFOLD} & \textbf{AFL} & \textbf{FedRS}  & \textcolor{gray}{\textbf{Oracle}} \\
\midrule
\multirow{2}{*}{Fashion-MNIST} & $C=3$ &$10$& $\mathbf{92.4 \pm 2.1}$ & $67.1 \pm 22.0$ & $66.9 \pm 20.8$ & $69.5 \pm 19.3$ & $78.9 \pm 14.7$ & $85.3 \pm 13.5$& \textcolor{gray}{$97.6 \pm 2.1$} \\
& $C=2$ & &$\mathbf{80.6 \pm 23.7}$ & $53.9 \pm 36.2$ & $52.9 \pm 35.7$ & $54.9 \pm 36.8$ & $78.6 \pm 20.0$ & $ 63.14 \pm 20.2$& \textcolor{gray}{$97.5 \pm 4.0$} \\
\midrule
\multirow{2}{*}{CIFAR-10} & $C=3$ && $\mathbf{65.6 \pm 10.1}$ & $44.0 \pm 8.4$ & $43.5 \pm 7.2$ & $43.3 \pm 7.4$ & $53.2 \pm 0.9$ & $44.0\pm 8.0$&  \textcolor{gray}{$85.5 \pm 5.0$} \\
& $C=2$ &$10$ & $\mathbf{72.8 \pm 17.4}$ & $46.7 \pm 15.8$ & $47.7 \pm 15.6$ & $46.7 \pm 14.9$ & $54.7 \pm 0.1$ & $49.4\pm 9.5$& \textcolor{gray}{$89.2 \pm 3.9$}\\
& $\beta=0.1$ & & $\mathbf{62.6 \pm 17.9}$ & $40.8	\pm 9.2$ & $41.9 \pm 9.7$ & $43.5 \pm 10.5$ & $53.4 \pm 11.5$ & $57.1\pm11.2$ &\textcolor{gray}{$79.2 \pm 3.7$}\\
\midrule
PACS & $C=6$ &$3$& $\mathbf{86.0\pm2.9}$ & $73.4\pm1.6$ & $75.3\pm1.3$& $73.9\pm0.3$ & $74.5 \pm 0.9$& $76.1 \pm 1.6$ &\textcolor{gray}{$90.5\pm 0.3$}\\
\bottomrule
\end{tabular}%
}
\label{tab:results}
\vspace{-4mm}
\end{table*}%
Dirichlet sampling simulates realistic non-i.i.d. label distributions by, for each client $i$, drawing a sample $p_i \sim \mathrm{Dirichlet}_K(\beta)$, where $p_i(k)$ represents the proportion of samples in client $i$ that have label $k \in [K]$. We use a symmetric concentration parameter $\beta > 0$ which controls the sparsity of the client distributions. See Appendix \ref{app:sample}.

While prior works have focused on inter-client distribution shifts assuming that client and target domains are equally distributed, \textit{we apply these sampling strategies also to the target set}, thereby introducing label shift between the client and target data. Figures~\ref{fig:label_sparsity_lab} \& \ref{fig:label_sparsity_dir} (latter in appendix) illustrate an example with $C=6$ for sparsity sampling and Dirichlet sampling with $\beta=0.1$, where the last client (Client 9) is chosen as the target. In addition, we investigate the effect of $T(Y)\notin\convh(S)$ in a task described in \ref{sec:synthexp}.

\paragraph{Baseline algorithms and model architectures} Alongside FedAvg, we use SCAFFOLD, FedProx, AFL and FedRS~\citep{scaffold,fedprox,mohri19a,li2021fedrs} as baselines. The first two chosen due to their prominence in the literature for handling non-iid data, and AFL which is similar in concept to \algname{} and aims to optimize for an unseen domain. We also include FedRS, designed specifically to address label distribution skew. 
For the synthetic experiment in Section~\ref{sec:synthexp}, we use a logistic regression model. For CIFAR-10 and Fashion-MNIST, we use small, two-layer convolutional networks, while for PACS and iWildCAM, we use a ResNet-50 pre-trained on ImageNet. Early stopping, model hyperparameters, and $\lambda$ in \algname{} are tuned using a validation set that reflects the target distribution in the synthetic experiment, CIFAR-10, Fashion-MNIST, and PACS. This tuning process consistently resulted in setting the number of local epochs to $E=1$ across all experiments. For iWildCAM, we adopt the hyperparameters reported by \citet{baibenchmarking} and select $\lambda$ using the same validation set used in their work. We report the mean test accuracy and standard deviation for each method over 3 independent random seeds for PACS and iWildCam and 8 seeds for the smaller Fashion-MNIST and CIFAR-10, to ensure robust evaluation. 


%
%
%
\subsection{Experimental results on benchmark tasks}

We present results for three tasks with selected skews in Table~\ref{tab:results} and explore detailed results below. Across these tasks, \algname{} consistently outperforms or matches the best-performing baseline. For PACS, Fashion-MNIST and CIFAR-10, we include results for an \textit{Oracle} FedAvg model, which is trained on clients whose distributions are identical to the target distribution, eliminating any client-target distribution shift (see Appendix \ref{app:expdetail} for details). A \algname{}-\textit{Oracle} would be equivalent since there is no label shift. The \textit{Oracle}, which has perfect alignment between client and target distributions, achieves superior performance, underscoring the challenge posed by distribution shifts in real-world scenarios where such alignment is absent.

\begin{figure*}[t]
    \centering
    \begin{subfigure}[b]{0.29\textwidth}
        \centering
    \includegraphics[width=1.\linewidth]{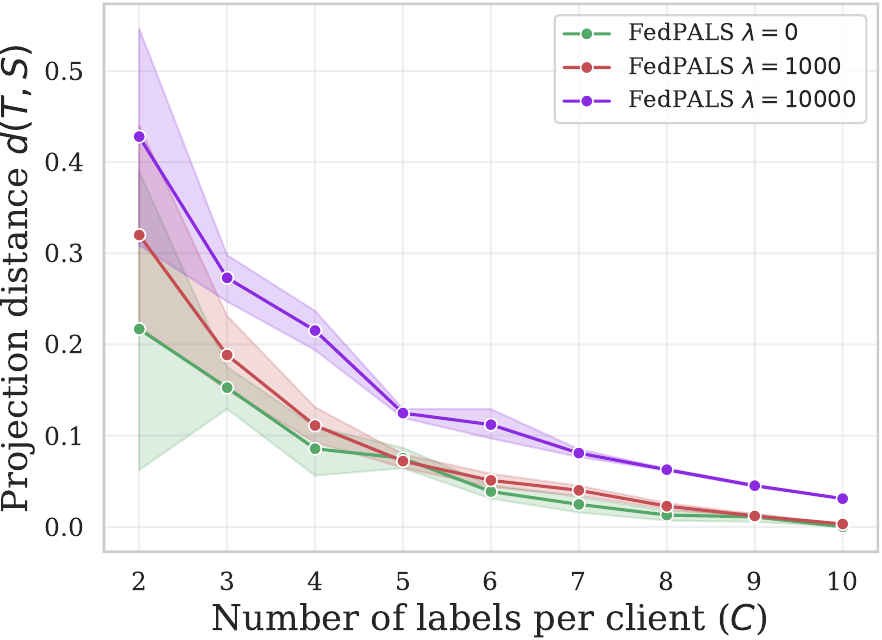}\vspace{1em}
        \caption{Projection distance between target and client convex hull, varying $C$.}
        \label{fig:projection_distance_sparsity}
    \end{subfigure}
    \;
    \begin{subfigure}[b]{0.29\textwidth}
        \centering
        \includegraphics[width=1.\linewidth]{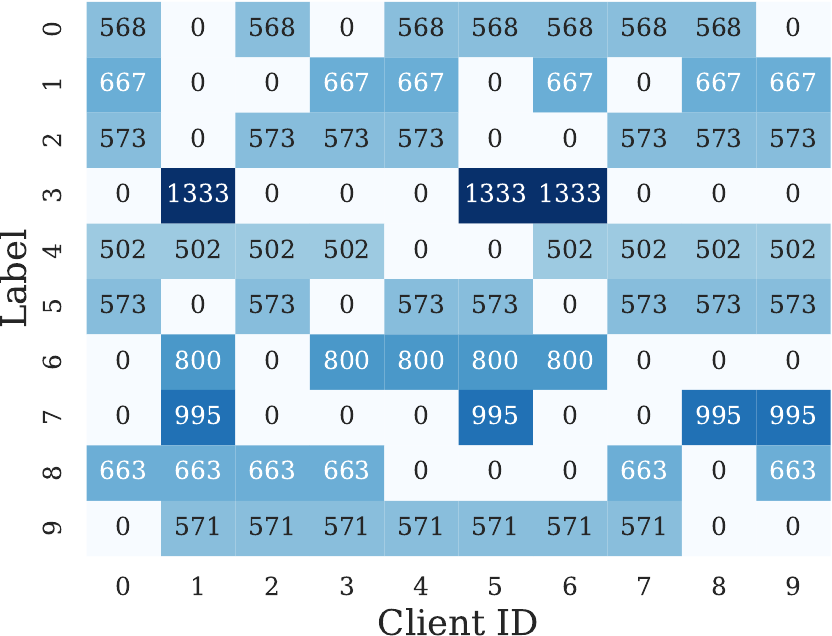}\vspace{1em}
        \caption{Marginal label distributions of clients and target with $C=6$ labels per client.}
        \label{fig:label_sparsity_lab}
    \end{subfigure}
    \;
    \begin{subfigure}[b]{0.36\linewidth}
        \centering
        \includegraphics[width=\linewidth]{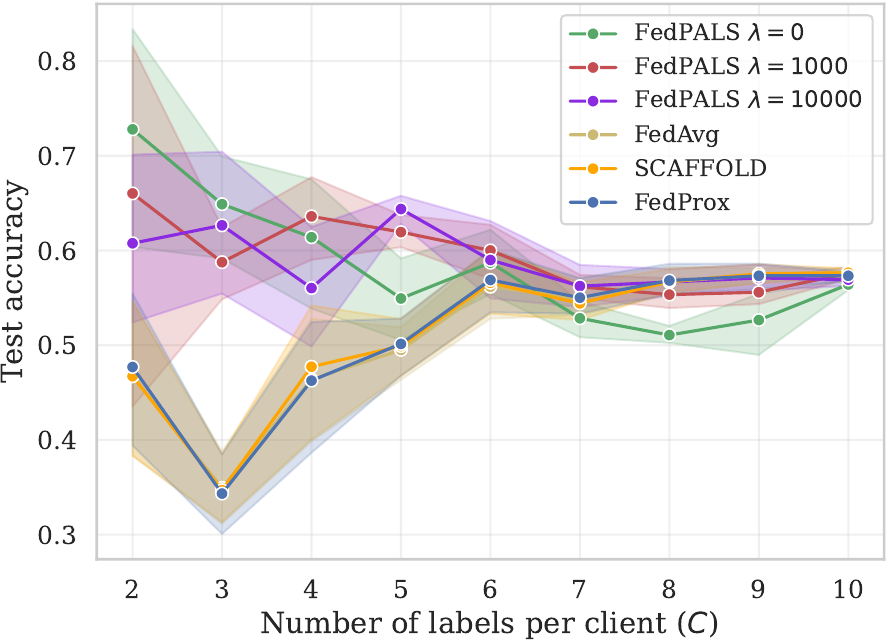}
        \caption{Accuracy vs labels per client, $C$.}
        \label{fig:fedpals_C}
    \end{subfigure}
    \quad
    \caption{Results on CIFAR-10 with sparsity sampling, varying the number of labels per clients $C$ across 10 clients. Clients with IDs 0--8 are used in training, and Client 9 is the target client. The task is more difficult for small $C$, when fewer clients share labels, and the projection distance is larger. }
    \label{fig:combined_sparsity}
\end{figure*}%
\begin{figure}[t!]
    \centering
    \begin{subfigure}{0.48\textwidth}
        \centering
        \includegraphics[width=.9\textwidth]{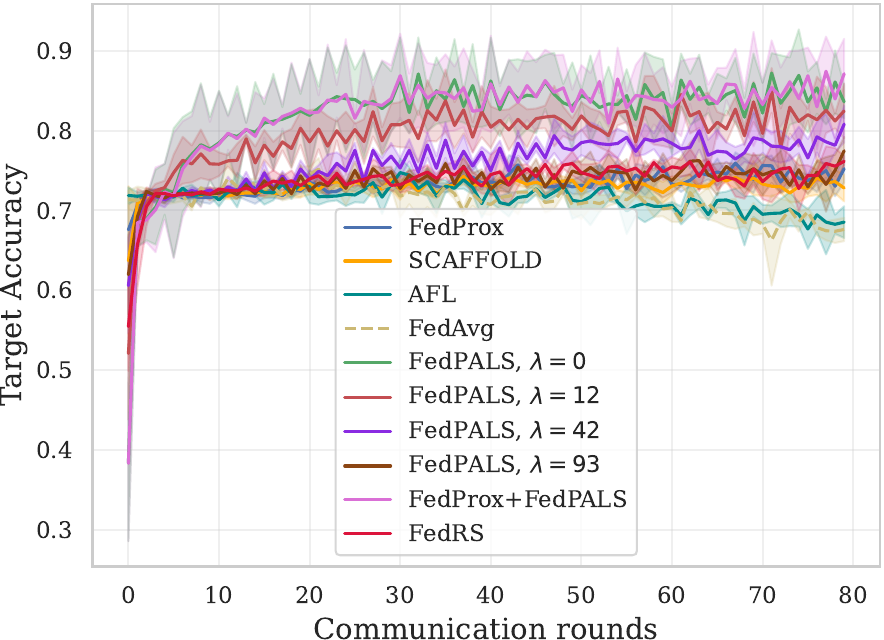}
        \caption{PACS, $M=3$.}
        \label{fig:pacspals}
    \end{subfigure}
    \begin{subfigure}{0.48\textwidth}
        \centering
        \includegraphics[width=.9\textwidth]{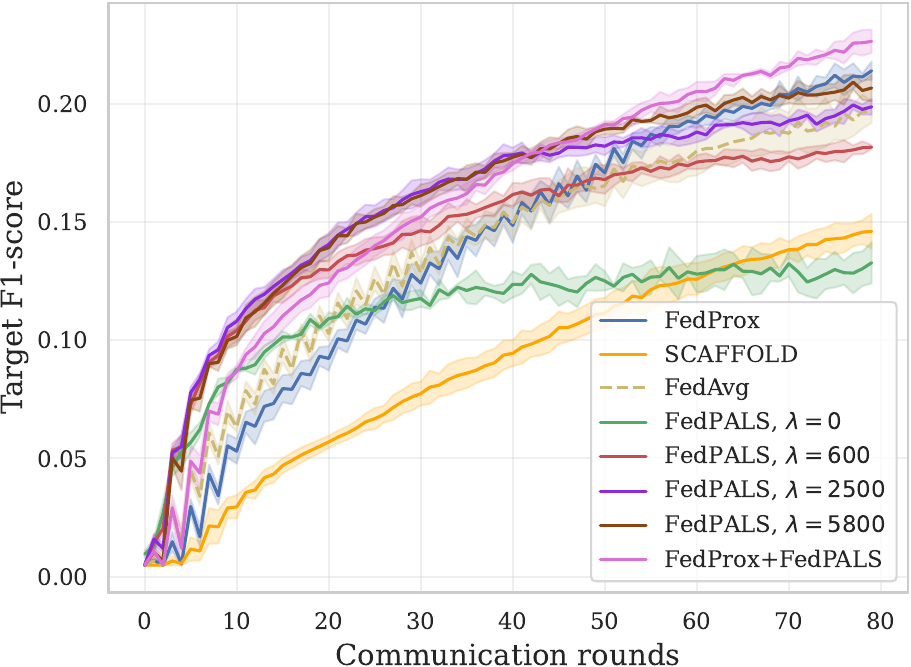}
          \caption{iWildCam, $M=100$.}
          \label{fig:wildcampals}
    \end{subfigure}
    \caption{Target accuracy/F1-score during training of \algname{} compared to baselines on PACS (a) and iWildCam (b),  averaged over 3 random seeds. $M$ is the number of training clients. Non-zero $\lambda$-values chosen to correspond to an ESS of $25\%$, $50\%$ and $75\%$.}
    \label{fig:dynamics}
    \vspace{-6mm}
\end{figure}%
\textbf{CIFAR-10/Fashion-MNIST.} Figure \ref{fig:fedpals_C} shows the results for the CIFAR-10 data set, where we vary the label sparsity across clients. In the standard i.i.d. setting, where all labels are present in both the training and target clients ($C=10$), all methods perform comparably. However, as label sparsity increases and fewer labels are available in client data sets (i.e., as $C$ decreases), we observe a performance degradation in standard baselines. In contrast, our proposed method, \algname{}, leverages optimized aggregation to achieve a lower target risk, resulting in improved test accuracy under these challenging conditions. Similar trends are observed for Fashion-MNIST, as shown in Figure \ref{fig:combined_C_fashion} in Appendix \ref{app:expresults}. Furthermore, the results in the highly non-i.i.d. cases ($C=2, 3$ and $\beta = 0.1$) are summarized in Table \ref{tab:results}. Additional experiments in Appendix \ref{app:expresults} examine how the algorithms perform with varying numbers of local epochs (up to 40) and clients (up to 100).

\textbf{PACS.}
As shown in Figure \ref{fig:pacspals}, being faithful to the target distribution is crucial for improved performance. Lower values of $\lambda$ generally correspond to better performance. Notably, FedAvg struggles in this setting because it systematically underweights the client with the distribution most similar to the target, leading to suboptimal model performance. In fact, this even causes performance to degrade over time. Interestingly, the baselines also face challenges on this task: both FedProx, FedRS and SCAFFOLD perform similarly to \algname{} when $\lambda=93$. However, \algname{} demonstrates significant improvements over these methods, highlighting the effectiveness of our aggregation scheme in enhancing performance. We also see that FedPALS + FedProx performs comparably to just using FedPALS in this case, although it does have higher variance. Additionally, in Table \ref{tab:results}, we present the models selected based on the source validation set, where \algname{} outperforms all other methods. For comprehensive results, including all \algname{} models and baseline comparisons, refer to Table \ref{tab:pacs} in Appendix \ref{app:expresults}.

\textbf{iWildCam.} The test performance across communication rounds is shown in Figure \ref{fig:wildcampals}. Initially, \algname{} widens the performance gap compared to FedAvg, but as training progresses, this gain diminishes. While \algname{} quickly reaches a strong performing model, it eventually plateaus. The rate of convergence and level of performance reached appears to be influenced by the choice of $\lambda$, with lower values of $\lambda$ leading to faster plateaus at lower levels compared to larger ones. This suggests that more uniform client weights and a larger effective sample size are preferable in this task. Given the iWildCam dataset's significant class imbalance -- with many classes having few samples -- de-emphasizing certain clients can degrade performance. We also note that our assumption of label shift need not hold in this experiment, as the cameras are in different locations, potentially leading to variations in the conditional distribution $p(X \mid Y)$. The performance of the models selected using the source validation set is shown in Table \ref{tab:wildcam} in Appendix \ref{app:expresults}. There we see that \algname{} performs comparably to FedAvg and FedProx while outperforming SCAFFOLD. Unlike in other tasks, where FedProx performs comparably or worse than \algname{}, we see FedProx achieve the highest F1-score on this task. Therefore, we conduct an additional experiment where we use both FedProx and \algname{} together, as they are not mutually exclusive. This results in the best performing model, see Figure \ref{fig:wildcampals}. Due to memory issues with the implementation FedRS was not able to run for this experiment and is omitted. AFL fails to learn in this task and is thus also omitted, although results are shown in Table \ref{tab:wildcam} in Appendix \ref{app:expresults}. Finally, as an illustration of the impact of increasing $\lambda$, we provide the weights of the clients in this experiment alongside the FedAvg weights in \ref{fig:alphas_ess} in Appendix~\ref{app:expresults}. We note that as $\lambda$ increases, the weights increasingly align with those of FedAvg while retaining weight on the clients whose label distributions most resemble that of the target.

%
%
\section{Discussion}
We have explored \textit{target-aware federated learning under label shift}, a scenario where client data distributions differ from a target domain with a known label distribution, but no target samples are available. We demonstrated that traditional approaches, such as federated averaging (FedAvg), which assume identical distributions between the client aggregate and the target, fail to adapt effectively in this context due to biased aggregation of client updates. To address this, we proposed \algname{}, a novel aggregation strategy that optimally combines client updates to align with the target distribution, ensuring that the aggregated model minimizes target risk. Empirically, across diverse tasks, we showed that under label shift, \algname{} significantly outperforms standard methods like FedAvg, FedProx, FedRS and SCAFFOLD, as well as AFL. Specifically, when the target label distribution lies within the convex hull of the client distributions, \algname{} finds the solution with the largest effective sample size, leading to a model that is most faithful to the target distribution. More generally, \algname{} balances the trade-off between matching the target label distribution and minimizing variance in the model updates. Our experiments further highlight that \algname{} excels in challenging scenarios where label sparsity and client heterogeneity hinder the performance of conventional federated learning methods.

One of the limitations of FedPALS is label shift--the assumption that label-conditional input distributions are equal in all clients and the target. We observed empirically that selecting the trade-off parameter $\lambda$ is crucial for optimal performance in tasks such as iWildCam, where this assumption may not fully hold. Future work should aim to detect and overcome violations of this assumption.

\clearpage

\bibliographystyle{plainnat}
\bibliography{references}

\clearpage
%
%
\appendix

%
%
\section{Experimental details}
\label{app:expdetail}
Here we provide additional details about the experimental setup for the different tasks. The code will be made available upon acceptance.
%
%
\subsection{Choice of hyperparameter $\lambda$}
\label{app:lambda}
A salient question in Cases 2 \& 3, as discussed in Section \ref{sec:limits}, is how to choose the strength of the regularization, $\lambda$. A larger value will generally favor influence from more clients, provided that they have sufficiently many samples. When $T \not\in \convh(S)$, the convex combination closest to $T$ could have weight on a single vertex. This will likely hurt the generalizability of the resulting classifier. In experiments, we compare values of $\lambda$ that yield different effective sample sizes, such as 10\%, 25\%, 50\% or 75\% of the original sample size, $N$. We can find these using binary search by solving \eqref{eq:alpha_alg} and calculate the ESS. One could select $\lambda$ heuristically based on the the ESS, or treat $\lambda$ as a hyperparameter and select it using a validation set. Although this would entail training and evaluating several models which can be seen as a limitation. We elect to choose a small set of $\lambda$ values based on the ESS heuristic and train models for these. Then we use a validation set to select the best performing model. This highlights the usefulness of the ESS as a heuristic. If it is unclear which values to pick, one could elect for a simple strategy of taking the ESS of $\lambda=0$ and 100\% and taking $l$ equidistributed values in between the two extremes, for some small integer $l$.
\subsection{Sampling strategies}
\label{app:sample}
Sparsity sampling entails randomly removing a subset of labels from clients and the target domain following the data set partitioning technique introduced in~\citet{mcmahan17a}. Each client is assigned $C$ random labels, with an equal number of samples for each label and no overlap among clients.
Sparsity sampling has been extensively used in subsequent studies~\citep{geyer2017differentially, li2020federated, li2022federated}.

Dirichlet sampling simulates non-i.i.d. label distributions by, for each client $i$, drawing a sample $p_i \sim \mathrm{Dirichlet}_K(\beta)$, where $p_i(k)$ represents the proportion of samples in client $i$ that have label $k \in [K]$. The concentration parameter $\beta > 0$ controls the sparsity of the client distributions.
In dirichlet sampling, using a smaller $\beta$ results in more heterogeneous client data sets, while a larger value approximates an i.i.d. setting. This widely-used method for sampling clients was first introduced by~\citet{yurochkin2019bayesian}. 
\subsection{Oracle construction}
The \textit{Oracle} method serves as a benchmark to illustrate the performance upper bound when there is no distribution shift between the clients and the target. To construct this \textit{Oracle}, we assume that the client label distributions are identical to the target label distribution, effectively eliminating the label shift that exists in real-world scenarios.

In practice, this means that for each dataset, the client data is drawn directly from the same distribution as the target. The aggregation process in the \textit{Oracle} method uses FedAvg, as no adjustments for label shift are needed. Since the client and target distributions are aligned, FedPALS would behave equivalently to FedAvg under this setting, as there is no need for reweighting the client updates.

This method allows us to assess the maximum possible performance that could be achieved if the distributional differences between clients and the target did not exist. By comparing the \textit{Oracle} results to those of our proposed method and other baselines, we can highlight the impact of label shift on model performance and validate the improvements brought by \algname{}.

\subsection{Perturbation of target marginal $\cT$}
In an experiment we perturb the given target label marginals, $\cT$, to evaluate the performance impact of noise in the estimate. We do this by generating gaussian noise, $\epsilon$, and then we add the noise to the label marginal to create a new target $\cT_p$. We modulate the size of the noise with a parameter $\delta$ and only add the positive noise values. 
$$
\cT_p=\cT + \delta\max (\epsilon,0)
$$
This is then normalised and used as the new target label marginal. 
This perturbation was done on the PACS experiment with $\delta\in [10^{-3},10^{-2},5\times10^{-1}]$ and repeated for three seeds. The results are given in Table \ref{tab:perturb} where we see that the performance decreases with increasing noise.
\begin{table}[]
    \centering
    \begin{tabular}{c|c}
    \hline
        $\delta$ & Accuracy \\
        \hline
        $10^{-3}$ & 88.8\\
        $10^{-2}$ & 85.2\\
        $5\times10^{-1}$ & 81.4\\
        \hline
    \end{tabular}
    \caption{Results of perturbing $\cT$ with varying noise levels $\delta$.}
    \label{tab:perturb}
\end{table}

\subsection{Synthetic task}\label{app:synth}
We randomly sampled three means $\mu_1 = [6, 4.6], \mu_2 = [1.2, -1.6],$ and $\mu_3 = [4.6, -5.4]$ for each label cluster, respectively.
\subsection{PACS}
In this task we use the official source and target splits which are given in the work by \citet{baibenchmarking}. We construct the task such that the training data is randomly assigned among three clients, then we remove the samples of one label from each of the clients. This is chosen to be labels '0', '1' and '2'. Then the client that is missing the label '2' is reduced so that it is $10\%$ the amount of the original size. For the target we modify the given one by removing the samples with label '2', thereby making it more similar to the smaller client. To more accurately reflect the target distribution we modify the source domain validation set to also lack the samples with label '2'. This is reasonable since we assume that we have access to the target label distribution. 


We pick four values of $\lambda$, [0,12,42,93], which approximately correspond to an ESS of $15\%, 25\%, 50\%$ and $75\%$ respectively.
We use the same hyperparameters during training as \citet{baibenchmarking} report using in their paper. Furthermore, we use the cross entropy loss in this task.
\subsection{iWildCam}
We perform this experiment using the methodology described in \citet{baibenchmarking} with the heterogeneity set to the maximum setting, i.e., $\lambda=0$ in their construction.\footnote{Note that this is not the same $\lambda$ used in the trade-off in \algname{}.} We use the same hyperparameters which is used for FedAvg in the same work to train \algname{}. We perform 80 rounds of training and, we then select the best performing model based on held out validation performance and report the mean and standard deviation over three random seeds. This can be seen in Table \ref{tab:wildcam}. We pick four values of $\lambda$, [0,600,2500,5800], which approximately correspond to an ESS of $8\%, 25\%, 50\%$ and $75\%$ respectively. We use the cross entropy loss in this task.

Due to FedProx performing comparably to FedPALS on this task, in contrast with other experiments, we also perform an experiment where we do both FedProx and \algname{}. This is easily done as FedProx is a client side method while \algname{} is a weighting method applied at the server. This results in the best performing model.

We use the same hyperparameters during training as \citet{baibenchmarking} report using in their paper. However, we set the amount of communication rounds to 80.

\section{Additional empirical results}
\label{app:expresults}
\begin{figure*}[ht!]
\centering
\begin{subfigure}[b]{0.33\textwidth}
    \centering
    \includegraphics[width=.99\textwidth]{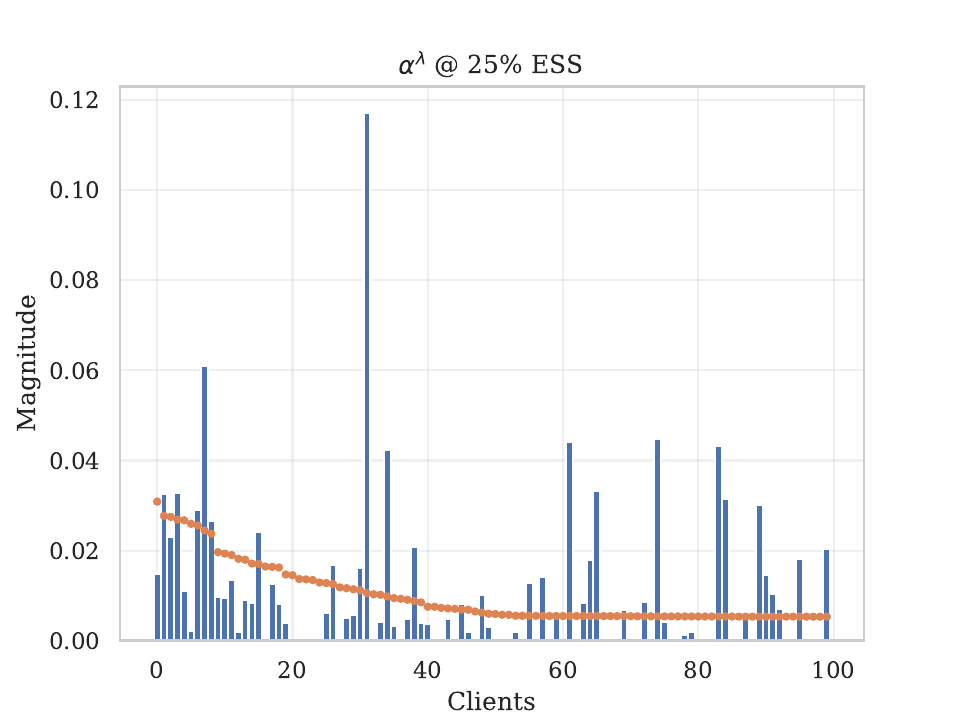}
    \caption{}
    \label{fig:ESS25}
\end{subfigure}%
\begin{subfigure}[b]{0.33\textwidth}
    \centering
    \includegraphics[width=.99\textwidth]{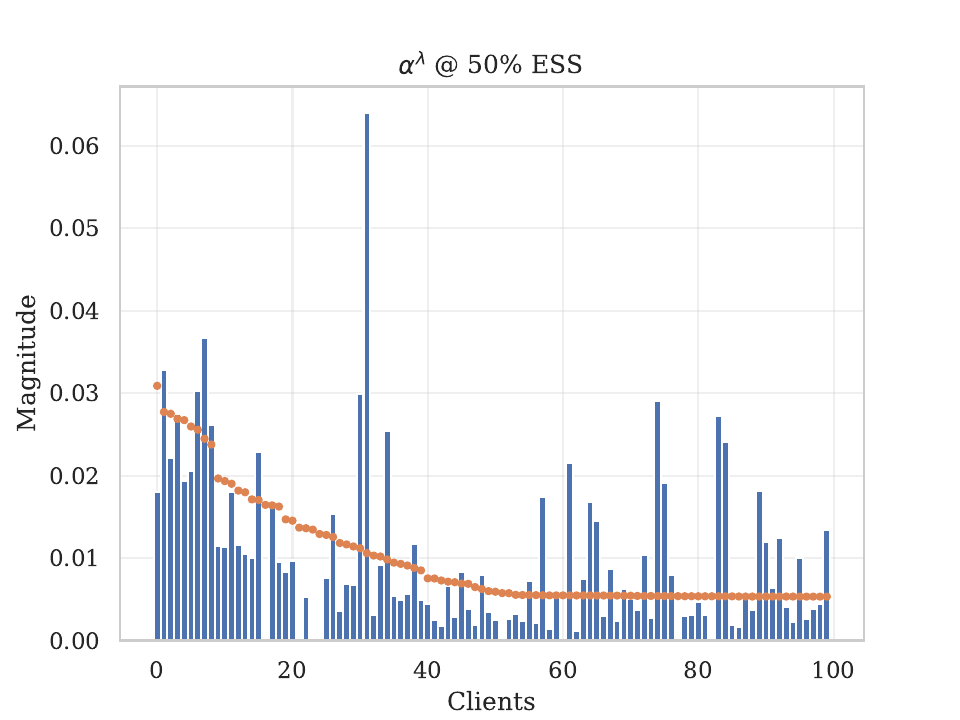}
    \caption{}
    \label{fig:ESS50}
\end{subfigure}%
\begin{subfigure}[b]{0.33\textwidth}
    \centering
    \includegraphics[width=.99\textwidth]{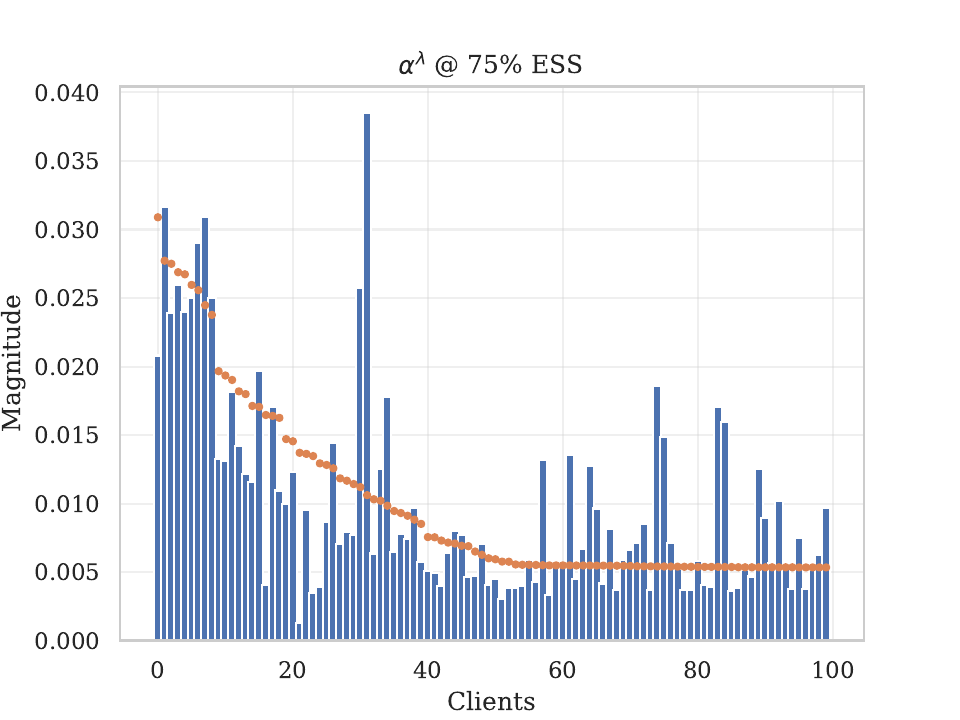}
    \caption{}
    \label{fig:ESS75}
\end{subfigure}%
\caption{An illustration of the aggregation weights of clients in the iWildCam experiment using \algname{} for different ESS. The clients are sorted by amount of samples in descending order. The magnitude of the weights produced by federated averaging is shown as dots. Note that with increasing the ESS, the magnitudes more closely resemble that of federated averaging.
\label{fig:alphas_ess}}
\end{figure*}

Figure~\ref{fig:alphas_ess} illustrates the aggregation weights of clients in the iWildCam experiment for $\lambda$ corresponding to different effective sample sizes.

We report the performance of the models selected using the held out validation set in Table \ref{tab:wildcam} and Table \ref{tab:pacs} for the iWildCam and PACS experiments respectively.

    \begin{table}[ht]
         \centering
    \caption{Results on iWildCam with 100 clients, standard deviation reported over 3 random seeds.}
    \begin{tabular}{lc}
    \toprule
    Algorithm & F1 (macro) \\
    \midrule
         \textbf{\algname, $\lambda=0$} & $0.13 \pm 0.00$ \\ 
         \textbf{\algname, $\lambda=600$} & $0.18\pm 0.00$\\
         \textbf{\algname, $\lambda=2500$} & $0.19\pm 0.00$\\ 
         \textbf{\algname, $\lambda=5800$} & $0.21\pm0.00$\\
         \textbf{FedProx+\algname, $\lambda=5800$} & $0.23\pm0.00$\\
         \textbf{FedAvg} & $0.20 \pm 0.01$\\
         \textbf{FedProx} & $0.21 \pm 0.00$\\
         \textbf{SCAFFOLD} & $0.15 \pm 0.01$\\
         \textbf{AFL} & $ 0.005\pm 0.0$ \\
         \bottomrule
    \end{tabular}    
    \label{tab:wildcam}
    \end{table}
\begin{table}[ht]
\centering
    \caption{Results on PACS with 3 clients with mean and standard deviation reported over 3 random seeds.}
    \begin{tabular}{lc}
    \toprule
    Algorithm & Accuracy \\
    \midrule
         \textbf{\algname, $\lambda=0$} & $86.0 \pm 2.9$ \\ 
         \textbf{\algname, $\lambda=12$} & $84.3\pm 2.5$\\
         \textbf{\algname, $\lambda=42$} & $81.7\pm 1.2$\\ 
         \textbf{\algname, $\lambda=93$} & $77.3\pm1.6$\\
          \textbf{FedProx+\algname, $\lambda=0$} & $87.2\pm4.1$\\
         \textbf{FedAvg} & $73.4\pm1.6$\\
         \textbf{FedProx} & $75.3\pm1.3$\\
         \textbf{SCAFFOLD} & $73.9\pm0.3$\\
         \textbf{AFL} & $ 74.5\pm 0.9$ \\
         \bottomrule
    \end{tabular}    
    \label{tab:pacs}
    \end{table}

\subsection{Results on CIFAR-10 with Dirichlet sampling}

\begin{figure}[ht]
    \centering
    \begin{subfigure}[b]{0.26\textwidth}
        \centering
        \includegraphics[width=1.\linewidth,height=1.1in]{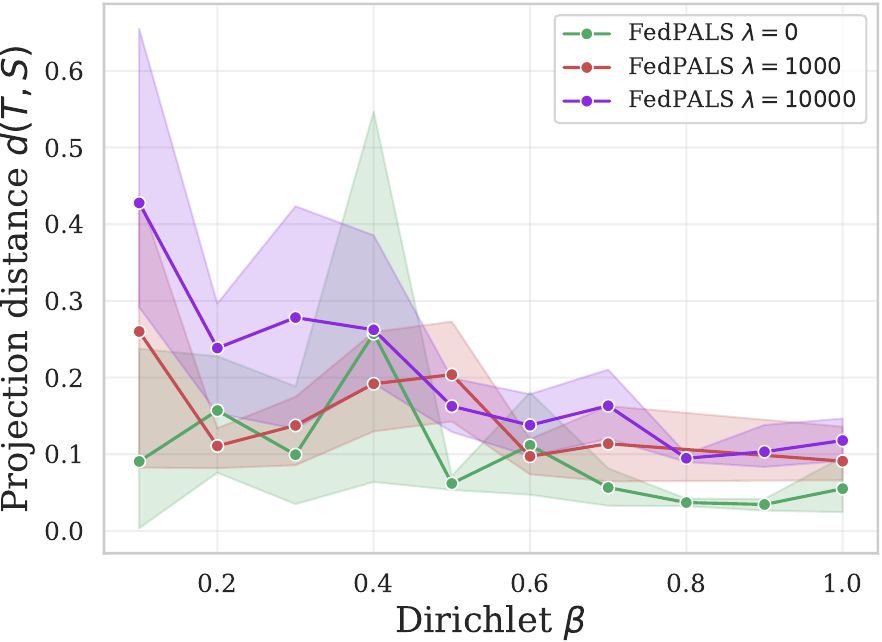}
        \caption{Projection distance between target label marginal and client convex hull}
        \label{fig:projection_distance_dirichlet}
    \end{subfigure}
    \;
    \begin{subfigure}[b]{0.26\textwidth}
        \centering
        \includegraphics[width=1.\linewidth]{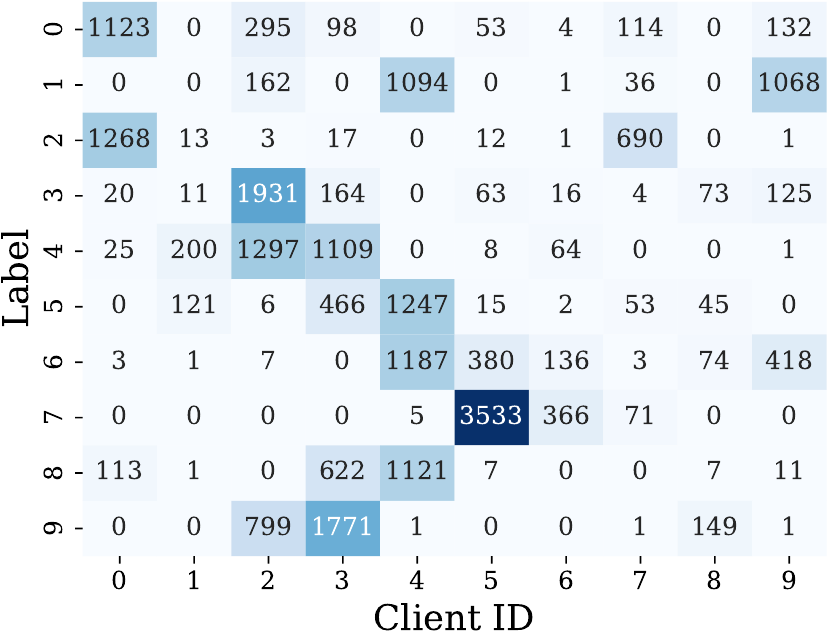}
        \caption{Marginal label distributions of clients and target with $\beta=0.1$.}
        \label{fig:label_sparsity_dir}
    \end{subfigure}
    \;
    \begin{subfigure}[b]{0.42\linewidth}
        \centering
        \includegraphics[width=\linewidth]{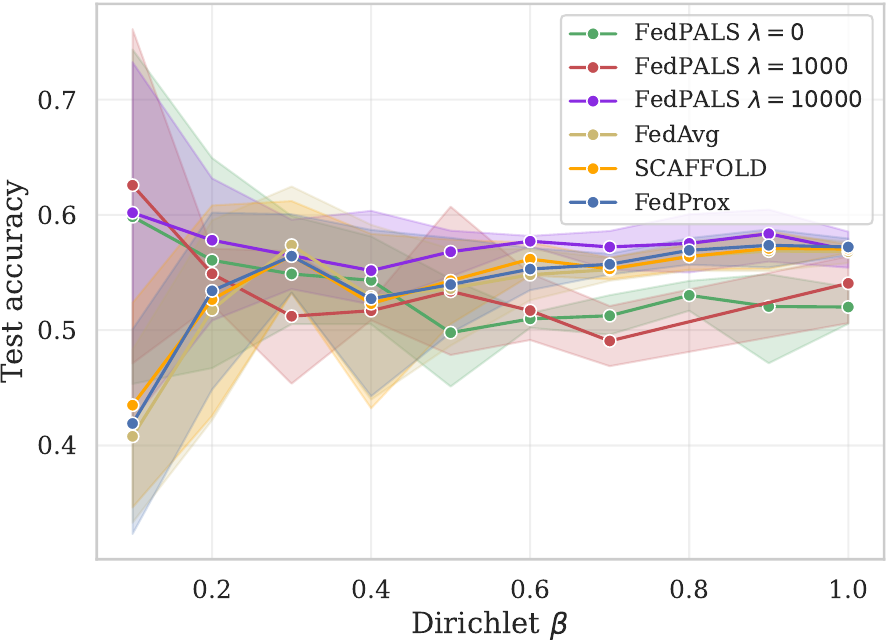}
        \caption{Accuracy under Dirichlet sampling.}
        \label{fig:dirichlet-sampling}
    \end{subfigure}
    \quad
    \caption{Results on CIFAR-10 with Dirichlet sampling across 10 clients, varying concentration parameter $\beta$. Clients with IDs 0--8 are clients present during training, and client with ID 9 is the target client.}
    \label{fig:combined_dirichlet}
\end{figure}

Figure~\ref{fig:combined_dirichlet} shows the results for the CIFAR-10 experiment with Dirichlet sampling of client and target label distributions.

\begin{figure}[ht]
    \centering
    \begin{subfigure}[b]{0.26\textwidth}
        \centering
        \includegraphics[width=1.\linewidth,height=1.1in]{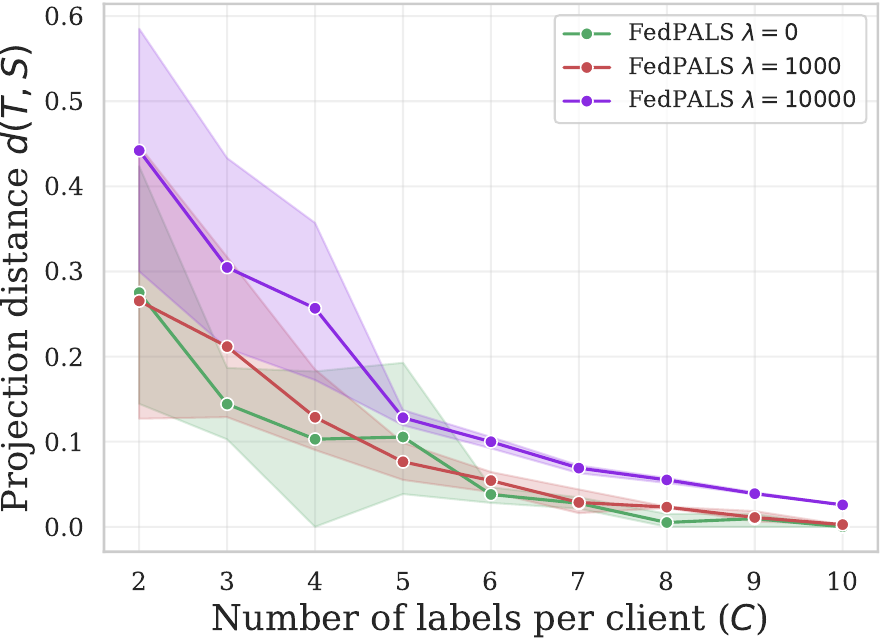}
        \caption{Projection distance between target label marginal and client convex hull}
        \label{fig:projection_distance_fmnist}
    \end{subfigure}
    \;
    \begin{subfigure}[b]{0.42\linewidth}
        \centering
        \includegraphics[width=\linewidth]{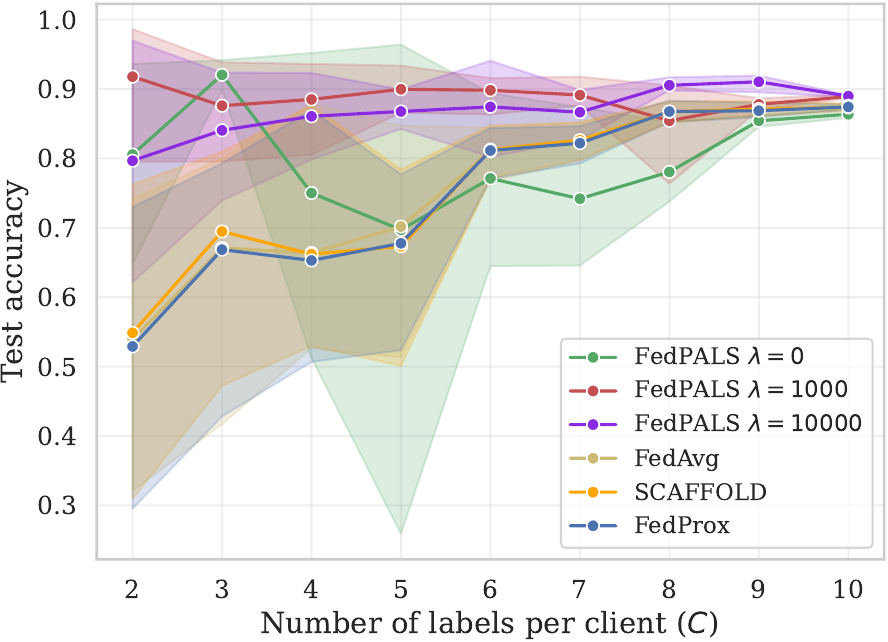}
        \caption{Accuracy under label sampling.}
        \label{fig:dirichlet-sampling-fmnist}
    \end{subfigure}
    \quad
    \caption{Results on Fashion-MNIST with label sampling across 10 clients, varying parameter $C$. Clients with IDs 0--8 are clients present during training, and client with ID 9 is the target client.}
    \label{fig:combined_C_fashion}
\end{figure}%

\subsection{Training dynamics for Fashion-MNIST}
\begin{figure}[ht]
    \centering
    \begin{subfigure}[b]{0.35\linewidth}
        \centering
        \includegraphics[width=\linewidth]{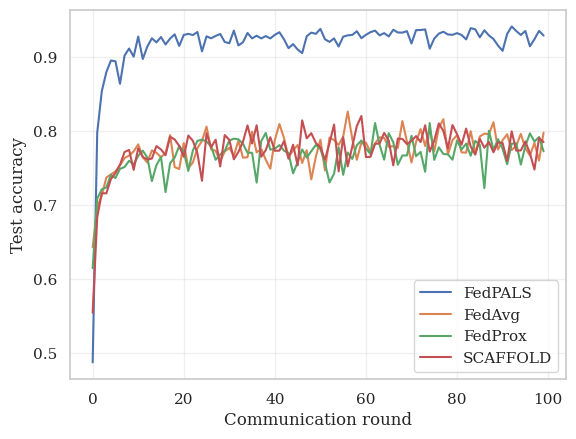}
        \caption{Fashion-MNIST with $C=6$.}
        \label{fig:testacc1}
    \end{subfigure}
    \hspace{0.05\linewidth}
    \begin{subfigure}[b]{0.35\linewidth}
        \centering
        \includegraphics[width=\linewidth]{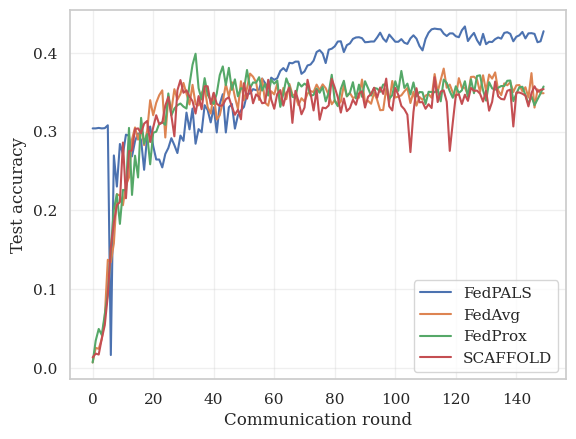}
        \caption{CIFAR-10 with Dirichlet $\beta=0.2$.}
        \label{fig:testacc2}
    \end{subfigure}
    \caption{Test accuracy during training rounds.}
    \label{fig:main-testacc}
\end{figure}

Figure~\ref{fig:main-testacc} shows the training dynamics for Fashion-MNIST and CIFAR-10 with different label marginal mechanisms.

\subsection{Local epochs and number of clients}
\begin{figure}[ht]
    \centering
    \begin{subfigure}[b]{0.32\linewidth}
        \centering
        \includegraphics[width=\linewidth]{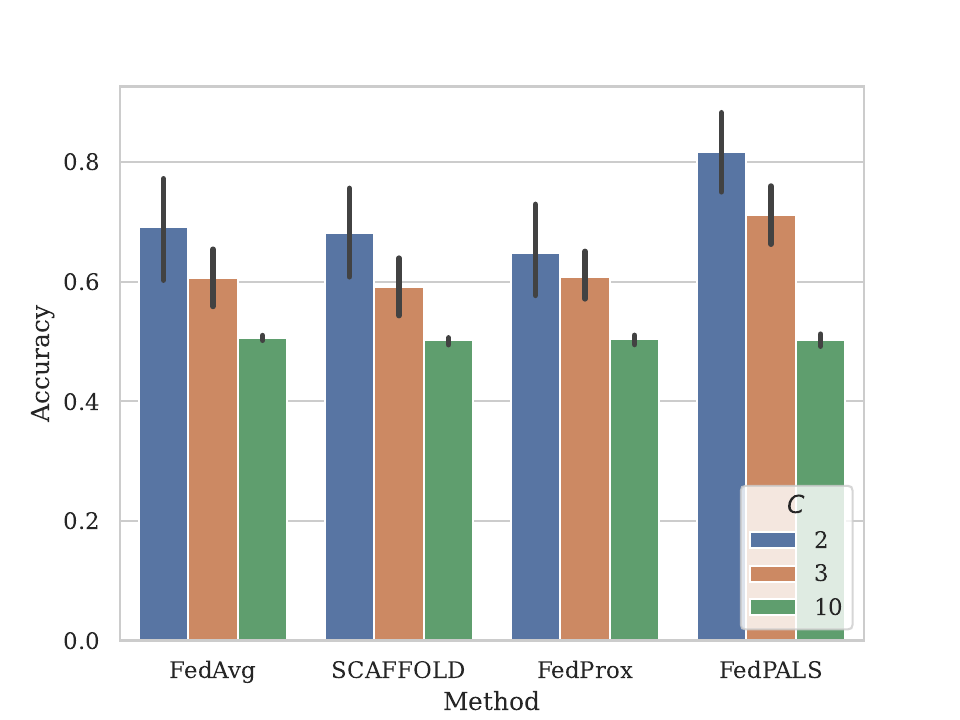}
        \caption{}
        \label{fig:100_clients}
    \end{subfigure}
    \begin{subfigure}[b]{0.32\linewidth}
        \centering
        \includegraphics[width=\linewidth]{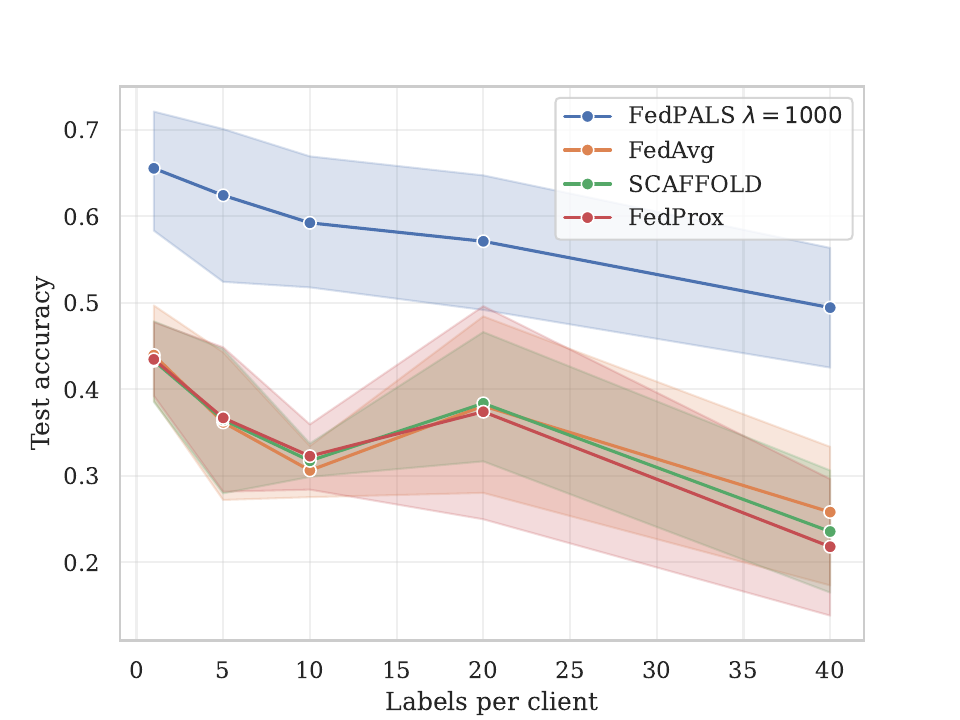}
        \caption{}
        \label{fig:local_ep}
    \end{subfigure}
    \begin{subfigure}[b]{0.32\linewidth}
    \centering
        \includegraphics[width=\linewidth]{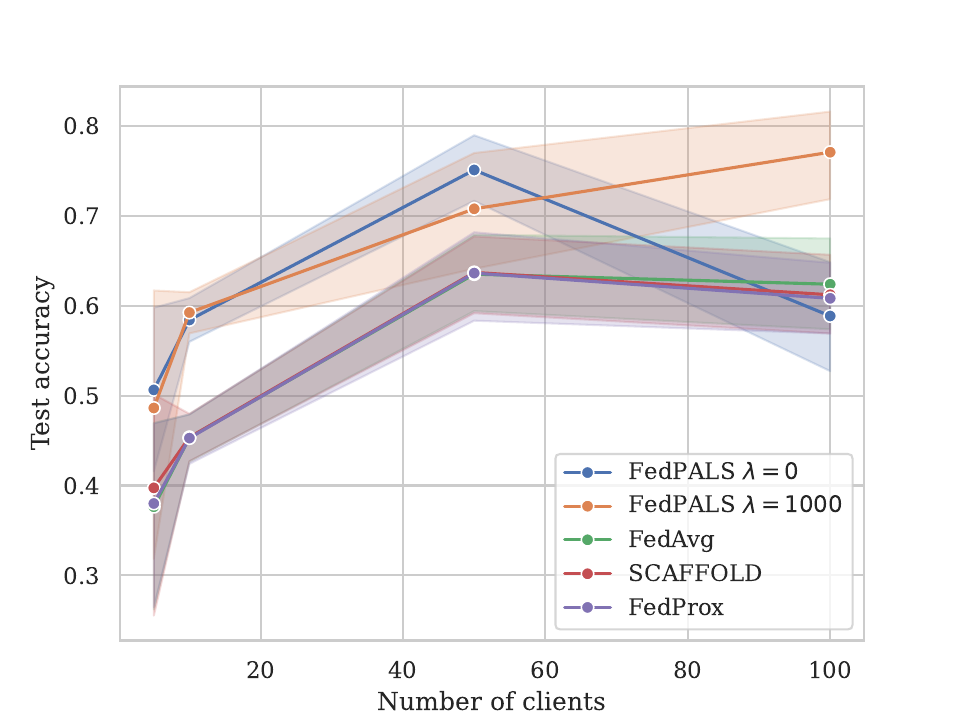}
        \caption{}
        \label{fig:n_clients}
    \end{subfigure}
    \caption{Comparison of CIFAR-10 results with different clients and settings. (a) 100 clients for $C=2,3,10$, $\lambda=1000$. (b) 10 clients and number of labels $C=3$. We plot test accuracy as a function of number of local epochs $E$. The total number of communication rounds $T$ are set such that $T = E/150$, where $150$ is the number of rounds used for $E=1$. (c) Test accuracy as a function of number of clients, with $C=3$.}
    \label{fig:combined-fig}
\end{figure}

In Figure \ref{fig:n_clients} we show results for varying number of clients for each method. For the cases with number of clients 50 and 100, we use the standard sampling method of federated learning where a fraction of 0.1 clients are sampled in each communication round. In this case, we optimize $\alpha^\lambda$  for the participating clients in each communication round. Interestingly, we observe that while FedAvg performs significantly worse than \algname{} on a target client under label shift, it outperforms both FedProx and SCAFFOLD when the number of local epochs is high ($E=40$), as shown in Figure \ref{fig:local_ep}.

%
%
\subsection{Synthetic experiment: effect of projection distance on test error}
\label{sec:synthexp}
\begin{figure}[t]
\centering
\includegraphics[width=.65\linewidth]{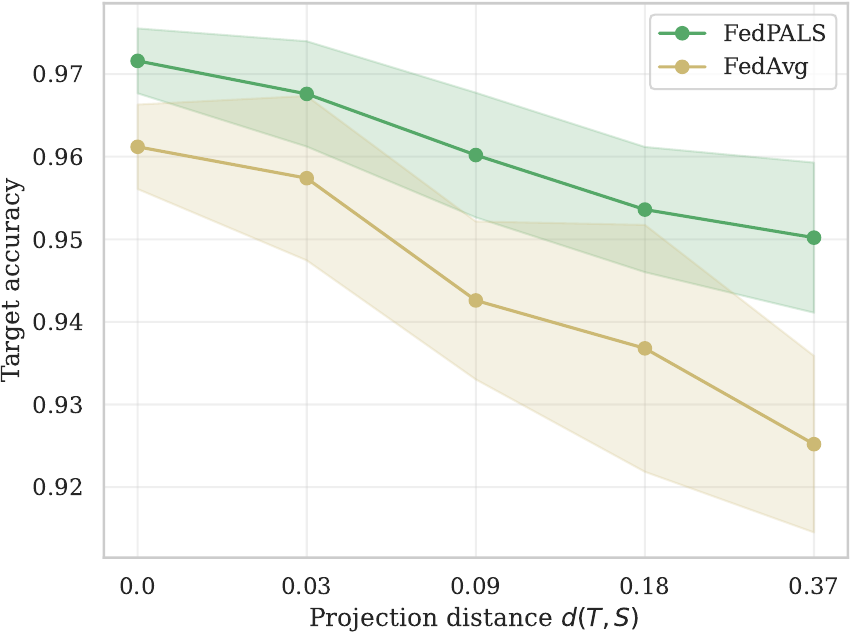}
\caption{Synthetic experiment. Accuracy of the global model as a function of the projection distance $d(T,S)$ between the target distribution $T(Y)$ and client label distributions $\convh(S(Y))$. Means and standard deviations reported over 5 independent runs.}\label{fig:projdist}
\end{figure}

When the target distribution $T(Y)$ is not covered by the clients, \algname{} finds aggregation weights corresponding to a regularized projection of $T$ onto $\convh(S)$. To study the impact of this, we designed a controlled experiment where the distance of the projection is varied. We create a classification task with three classes, $\cY=\{0, 1, 2\}$, and define $p(X \mid Y=y)$ for each label $y \in \cY$ by a unit-variance Gaussian distribution $\mathcal{N}(\mu_y, I)$, with randomly sampled means $\mu_y \in \mathbb{R}^2$. We simulate two clients with label distributions $S_1(Y)=[0.5,0.5,0.0]^\top$ and $S_2(Y)=[0.5,0.0,0.5]^\top$, and $n_1=40$, $n_2=18$ samples, respectively. Thus, FedAvg gives larger weight to Client 1. 
We define a target label distribution $T(Y)$ parameterized by $\delta\in[0,1]$ which controls the projection distance $d(T,S)$ between $T(Y)$ and $\convh(S)$, 
\begin{equation*}
    T_\delta(Y) \coloneqq (1-\delta)T_{\text{proj}}(Y) + \delta T_{\text{ext}}(Y)~,
\end{equation*}
with $T_{\text{ext}}(Y) = [0,0.5,0.5]^\top \notin \convh(S(Y))$ and $T_{\text{proj}}(Y) = [0.5, 0.25, 0.25]^\top \in\convh(S(Y))$. By varying $\delta$, we control the projection distance $d(T,S)$ between each $T_\delta$ and $\convh(S)$ from solving \eqref{eq:alpha0}, allowing us to study its effect on model performance.

We evaluate the global model on a test set with $n_{\text{test}}=2000$ samples drawn from the target distribution $T(Y)$ for each value of $\delta$ and record the target accuracy for \algname{} and FedAvg. 
Figure \ref{fig:projdist} illustrates the relationship between the target accuracy and the projection distance $d(T,S)$ due to varying $\delta$. When $d(S,T)=0$ (i.e., $T(Y)\in\convh(S)$), the target accuracy is highest, indicating that our method successfully matches the target distribution. As $d(S,T)$ increases (i.e., $T$ moves further away from $\convh(S)$), the task becomes harder and accuracy declines. For all values, \algname{} performs better than FedAvg. For more details on the synthetic experiment, see Appendix~\ref{app:expdetail}.

%
%
\section{Proofs}
\label{app:proofs}

\subsection{\algname{} updates}

\begin{repprop}{prop:sgd_update}[Unbiased SGD update]
    Consider a single round $t$ of federated learning in the batch stochastic gradient setting with learning rate $\eta$. Each client $i\in [M]$ is given parameters $\theta_t$ by the server, computes their local gradient, and returns the update $\theta_{i, t} = \theta_t - \eta \nabla_\theta \hat{R}_i(h_{\theta_t})$. Let weights $\alpha^c$ satisfy $T(X,Y) = \sum_{i=1}^M\alpha^c_i S_i(X,Y)$. Then, the aggregate update $\theta_{t+1} = \sum_{i=1}^M \alpha^c_i \theta_{i,t}$ satisfies
    $$
    \E[\theta_{t+1} \mid \theta_t] = \E[\theta_{t+1}^T \mid \theta_t]~, 
    $$
    where $\theta_{t+1}^T$ is the batch stochastic gradient update for $\hat{R}_T$ that would be obtained with a sample from the target domain.  
\end{repprop}
\begin{proof}
\begin{align}
    \theta_{t+1} & = \sum_{i=1}^M \alpha^c_i \theta_{i, t} = \sum_{i=1}^M \theta^c_i(\theta_t - \eta\nabla \hat{R}_i(h_{\theta_t})) 
     = \theta_t - \eta \sum_{i=1}^M \alpha_i \nabla \hat{R}_i(h_{\theta_t})
\end{align}
\begin{align}
    \E[\theta_{t+1} \mid \theta_t] &= \theta_t - \eta \cdot  \E\left[\sum_{i=1}^M \alpha_i \nabla \hat{R}_i(h_{\theta_t}) \mid \theta_t \right] \\
    & = \theta_t - \eta \cdot \sum_{x,y }\E\left[\sum_{i=1}^M \hat{S}_i(x,y) \alpha_i \right] \nabla L(y, h_{\theta_t}(x)) \\
    & = \theta_t - \eta \cdot \sum_{x,y }T(x, y) \nabla L(y, h_{\theta_t}(x))  \\
    & = \theta_t - \eta \cdot \E\left[\sum_{x,y }\hat{T}(x, y)\right] \nabla L(y, h_{\theta_t}(x)) = \E[\theta^T_{t+1} \mid \theta_t]~.
\end{align}
\end{proof}

\subsection{\algname{} in the limits}
As $\lambda\rightarrow \infty$, because the first term in \eqref{eq:alpha_alg} is bounded, the problem shares solution with 
\begin{equation}\label{eq:lambdainfprob}
\min_{\alpha_1, \ldots, \alpha_M} \sum_i \frac{\alpha_i^2}{n_i} \quad \mbox{s.t.} \quad \sum_i \alpha_i=1, \quad \forall i : \alpha_i\geq 0~.
\end{equation}
Moreover, we have the following result.

\begin{thmprop}\label{prop:lambdainf}
    The optimization problem
    $$
    \min_\alpha \sum_i \frac{\alpha_i^2}{n_i} \quad s.t \quad \sum_i \alpha_i=1 \quad \alpha_i\geq 0 ~ \forall~i~,
    $$
    has the optimal solution $\alpha_i^*=\frac{n_i}{\sum_i n_i}$ where $i\in[1,m]$
\end{thmprop}
\proof
From the constrained optimization problem we form a Lagrangian formulation
$$
\cL(\alpha, \mu, \tau) =  \sum_i \frac{\alpha_i^2}{n_i} + \mu \underbrace{(1-\sum_i \alpha_i)}_{h(\alpha)} + \tau\underbrace{-\alpha}_{g(\alpha)}
$$
We then use the KKT-theorem to find the optimal solution to the problem. 
\begin{equation}
    \nabla_\alpha \cL (\alpha^*) =0 \implies \forall i :  ~~2\frac{\alpha^*_i}{n_i} -\mu - \tau =0 \label{eq:grad}~.
\end{equation}
In other words, the following ratio is a constant, 
$$
\forall i \quad \frac{\alpha^*_i}{n_i} = c
$$
for some constant $c$.
We have the additional conditions of primal feasibility, i.e.
\begin{align*}
    & h(\alpha^*)=0 \\
    & g(\alpha^*)\leq 0 
\end{align*}
From the first constraint, we have $\sum_{i=1}^M \alpha^*_i = 1$, and thus, 
$$
\sum_{i=1}^M \alpha_i^* = c \sum_{i=1}^M n_i = 1
$$
which implies that $c = 1/\sum_{i=1}^M n_i$ and thus
$$
\forall i : \alpha_i^* = \frac{n_i}{\sum_{i=1}^M n_i}~.
$$

\endproof

\end{document}

%% file: header.tex
\def\algname{FedPALS}
\def\algfullname{Federated learning with Parameter Aggregation for Label Shift}

 \DeclareMathOperator*{\argmin}{arg\,min}

\newtheorem{thmprop}{Proposition}

\newtheorem{thmasmp}{Assumption}
\newtheorem*{thmidasmp*}{Identifying assumptions}

\theoremstyle{definition}

\newtheorem*{thmrem*}{Remark}
\newtheorem*{thmprop*}{Proposition}

\makeatletter
\newtheorem*{rep@theorem}{\rep@title}
\newcommand{\newreptheorem}[2]{%
\newenvironment{rep#1}[1]{%
 \def\rep@title{#2 \ref{##1} (Repeated)}%
 \begin{rep@theorem}}%
 {\end{rep@theorem}}}
\makeatother

\newreptheorem{theorem}{Theorem}
\newreptheorem{lemma}{Lemma}
\newreptheorem{col}{Corollary}
\newreptheorem{def}{Definition}
\newreptheorem{prop}{Proposition}

\def\convh{\mathrm{Conv}}

\def\E{\mathbb{E}}

\def\bbR{\mathbb{R}}

\def\cL{\mathcal{L}}

\def\cT{\mathcal{T}}

\def\cX{\mathcal{X}}
\def\cY{\mathcal{Y}}